\documentclass[letterpaper, 10 pt, journal, twoside]{./style/IEEEtran}
 
\usepackage[pdftex]{graphicx}
\graphicspath{{images-src/}{images-bin/}} % Read from both source and compiled image directories. 

\usepackage{amssymb}
\usepackage{amsmath}
\usepackage{cleveref}
\usepackage[caption=false,font=normalsize,labelfont=sf,textfont=sf]{subfig}
\usepackage{graphicx}
\usepackage{multirow}
\usepackage{array}

\Crefname{equation}{Eq.}{Eqs.} % Ensure abbreviation used for capital version also.
\Crefname{figure}{Fig.}{Figs.}

\usepackage[textsize=scriptsize]{todonotes}
\newcommand{\newtext}[1]{{\color{blue}{#1}}}

\IEEEoverridecommandlockouts                             
\title{Bayesian optimization for robust robotic grasping using a sensorized compliant hand}

\author{Juan G. Lechuz-Sierra$^1$, Ana Elvira H. Martin$^2$, Ashok M. Sundaram$^2$,\\Ruben Martinez-Cantin$^1$, M\'{a}ximo A. Roa$^2$
\thanks{Manuscript received: May, 27, 2024; Revised July, 27, 2024; Accepted September, 19, 2024.}%Use only for final RAL version
\thanks{This paper was recommended for publication by Editor Editor T. Ogata upon evaluation of the Associate Editor and Reviewers' comments.} %Use only for final RAL version
\thanks{$^1$ Instituto de Investigación en Ingeniería de Aragón (I3A) and DIIS, Universidad de Zaragoza, Spain}
\thanks{$^2$ German Aerospace Center (DLR), Institute of Robotics and Mechatronics (RM), M\"unchener Str. 20, 82234 We\ss ling, Germany}
\thanks{Digital Object Identifier (DOI): see top of this page.}

}
\begin{document}

\maketitle

%beginispell

%%%%%%%%%%%%%%%%%%%%%%%%%%%%%%%%%%%%%%%%%%%%%%%%%%%%%%%%%%%%%%%%%%%%%%%%%%%%%%%%
\begin{abstract}
One of the first tasks we learn as children is to grasp objects based on our tactile perception. Incorporating such skill in robots will enable multiple applications, such as increasing flexibility in industrial processes or providing assistance to people with physical disabilities. However, the difficulty lies in adapting the grasping strategies to a large variety of tasks and objects, which can often be unknown. The brute-force solution is to learn new grasps by trial and error, which is inefficient and ineffective. In contrast, Bayesian optimization applies active learning by adding information to the approximation of an optimal grasp. This paper proposes the use of Bayesian optimization techniques to safely perform  robotic grasping. We analyze different grasp metrics to provide realistic grasp optimization in a real system including tactile sensors. An experimental evaluation in the robotic system shows the usefulness of the method for performing unknown object grasping even in the presence of noise and uncertainty inherent to a real-world environment.
\end{abstract}


%%%%%%%%%%%%%%%%%%%%%%%%%%%%%%%%%%%%%%%%%%%%%%%%%%%%%%%%%%%%%%%%%%%%%%%%%%%%%%%%
\section{Introduction}

\IEEEPARstart{E}{arly} work on robotic grasp prediction, including the work carried out by Miller et al. in \cite{miller2003automatic}, was based on deterministic models with programmed grasping sequences constrained by the need of having a precise geometric and physical model of the object to be grasped. Later works have shifted the focus to more data-driven and adaptive grasping strategies, which are more flexible, robust and reliable, such as the approach by Yang et al. \cite{yang2023sim} or the Dex-Net system developed in \cite{mahler2017dex} by Mahler et al. However, these approaches are not without drawbacks since they are computationally expensive and require large amounts of data to be trained. On the other hand, to enhance the efficiency and adaptability of a grasping method it is crucial to make it transferable to any object, even without previous knowledge of its characteristics. A simple solution to address the grasping of an unknown object could be to learn new grasps by trial-and-error methods, which can be highly inefficient and computationally demanding. In this line, different model-free approaches like the work carried out by Mahler et al. \cite{mahler2015gp} and Yi et al. \cite{Yi2016} have been proposed, planning the grasp based on visual and tactile feedback and statistical foundations with greater interpretability than deep learning approaches. Among them, Bayesian Optimization (BO) \cite{brochu2010tutorial} has the capacity to leverage online grasp memory, making it a powerful tool for creating versatile robotic grasping systems. It is the most sample efficient global optimization method, reducing the number of trials required to find an optimal grasp, balancing between exploring new configurations and exploiting the learned probabilistic features that describe the process, while handling the uncertainty inherent in robotic perception and control. In particular, working with an approach based solely on tactile detection, with no external sensor for shape recognition, allows us to confront BO with the uncertainty introduced by realistic sensors, for analytically measuring the stability of the grasp. Nevertheless, its sample efficiency makes it suitable to deal with general black-box functions that may include visual feedback to ensure robustness.\\

\begin{figure}
    \centering
    \includegraphics[width=0.75\linewidth]{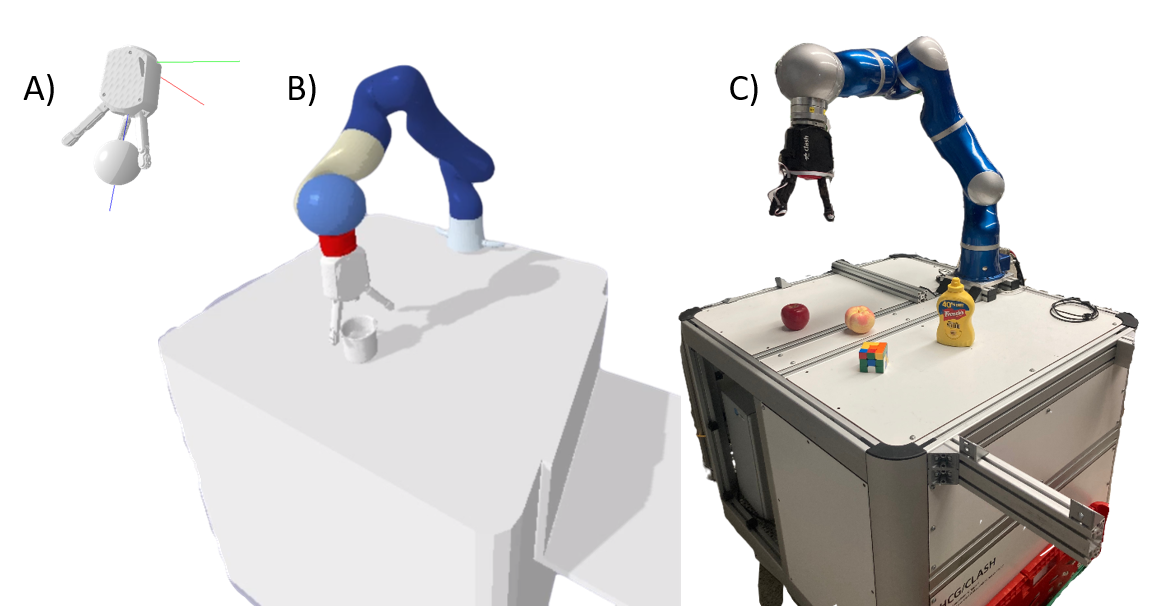}
    \caption{Different setups used during experimentation with Bayesian optimization-based grasps: A) Simulation scenario, optimizing the grasp by moving CLASH hand freely around the object; B) Simulation of the complete robot model, with motion and collision constraints due to the arm and workbench; C) Real experimentation environment, including CLASH, a compliant under-actuated hand with tactile sensors on the fingertips.}
    \label{fig:setups}
\end{figure}

In this work, a BO approach based on tactile feedback is used to handle complexity and uncertainty introduced by a real-world setup  (Fig. \ref{fig:setups}C), proving the generality of the method to obtain robust and safe grasps when dealing with diverse objects and end-effectors just by using the knowledge from previous trials. The specific contributions of this paper are:

\begin{itemize}
    \item An exploration system for robotic grasping in 3D space based on BO, applied to a robotic hand endowed with tactile sensors.
    \item The development of new heuristics that improve the convergence of the grasp evaluation function, increasing the applicability and task success in real environments.
    \item Verification of the portability of the method to a real system, evaluating grasp using contact points-based analytic metrics and real tactile sensor measurements.
\end{itemize}

%%%%%%%%%%%%%%%%%%%%%%%%%%%%%%%%%%%%%%%%%%%%%%%%%%%%%%%%%%%%%%%%%%%%%%%%%%%%%%%%
\section{Related work}
\textbf{Robotic grasping} has greatly evolved from only relying on pre-programmed sequences, such as the work by Miller et al. in \cite{miller2003automatic}, where the modeling of the object as a shape primitive allows the selection of a starting grasp position. To compensate the lack of precision and adaptability of such approaches, learning-based methods have lately received increased attention. Data-driven strategies have shown improved robustness through the use of extensive real-world data. This is the case of the Dex-Net system developed by Mahler et al. in \cite{mahler2017dex}, where sets of point-clouds were used to learn the properties of the object. Although that work was focused on reducing the required data to train, deep learning methods are still computationally expensive. To reduce this training time, Yan et al. \cite{yan2022robotic} proposed to use explicit ensemble methods. However, deep-learning-based methods fail to generalize to new objects different from those trained on. Model-free approaches have been explored to facilitate the grasping of previously unknown objects. They rely on exploring grasp possibilities and object shape, according to probabilistic foundations. In the case of Mahler et al. \cite{mahler2015gp}, grasp planning is addressed in 2D using visual data only. On the other hand, Yi et al. \cite{Yi2016} used tactile feedback to reduce the object surface uncertainty.

To improve the efficiency of exploration and grasp sampling, \textbf{Bayesian Optimization} has emerged as a promising technique with a sequential strategy for addressing systematic uncertainties while providing an effective solution in many robotics and reinforcement learning problems, such as robotic walking \cite{calandra2016bayesian}, adaptive control \cite{kuindersma2013variable}, and path planning \cite{martinez2009bayesian}. Its effectiveness in robotic grasping was first studied in the work of Daniel et al. \cite{daniel2015active}, where a human acted as an expert to evaluate grasps. Later, Nogueira et al. \cite{nogueira2016unscented} introduced the unscented transformation as a system to propagate the noise distribution through decisions, enabling the identification of optimal and safe grasps, with no human needed in the loop. The work was later continued by Castanheira et al. in \cite{castanheira2018finding}, where a haptic exploration strategy combining BO with a collision penalty heuristic is proposed, accelerating the method's convergence. Another important contribution was made by De Farias et al. in \cite{de2021simultaneous}, introducing object shape representation and tactile exploration into the grasp planning process, achieving stable grasps with a high probability of success. Recently, our contribution with Herrera et al. \cite{herrera2022optimizacion}, compared different parameterizations of the grasp to be used with BO, as well as BO algorithm variants to obtain multiple grasp solutions. Compared to previous work, this work confronts the probabilistic nature of BO directly with the uncertainty introduced by real tactile sensors, which makes contact points-based metrics less reliable.

\textbf{Tactile sensing} has been utilized in intelligent robotics for slip detection \cite{romeo2020methods}, manipulation \cite{yamaguchi2019recent} and object properties perception \cite{luo2017robotic}, emulating human capabilities by multi-modal signal measurement, such as in the sensors used in this work with tri-axis force measurements. In addition, a tridactyl under-actuated soft hand has been used, which, while increasing the grasping capabilities \cite{bonilla2014grasping}, decreases the measurement accuracy, due to the positional error introduced when computing the analytical metrics. This increases complexity compared to grasping with pincer-like grippers used in other grasping approaches so far.

%%%%%%%%%%%%%%%%%%%%%%%%%%%%%%%%%%%%%%%%%%%%%%%%%%%%%%%%%%%%%%%%%%%%%%%%%%%%%%%%
\section{Bayesian optimization}

This section explains the mathematical background allowing BO to efficiently tackle the global optimization of black-box functions, e.g. grasping an object. \newtext{More details about the notation, implementation and pseudo-code can be found in the literature \cite{brochu2010tutorial,martinez2014bayesopt}}.

Formally, BO finds the optimum of an unknown real-valued function $\mathnormal{f}: \mathcal{X} \rightarrow \mathbb{R}$, where  $\mathcal{X} \subset \mathbb{R}^{d}$ is a compact space, in our case corresponding to the space of parameters describing the end-effector pose. The algorithm selects the best query point $x_{i} \in \mathcal{X}$ to evaluate each iteration with outcome $y_{i} = \mathnormal{f}(x_{i}) + \eta$, where $\eta$ is zero-mean noise with variance $\sigma_{\eta}^{2}$, finding an optimum $x^{*}$ so that $|y^{*} - y_{n}|$ is minimum for the considered budget.

BO is composed of two main ingredients. The first one is a \textbf{probabilistic surrogate model}, which is a distribution over the family of functions $\mathnormal{P(f)}$, where $\mathnormal{f}$ belongs, built by incrementally sampling over $\mathcal{X}$. As implemented in BayesOpt \cite{martinez2014bayesopt} we consider the probabilistic model as a Gaussian process $GP(x|\mu, \sigma^{2}, \theta)$ with inputs $x \in \mathcal{X}$, scalar outputs $y \in \mathbb{R}$ and an associated kernel or covariance function $k$ with hyperparameters $\theta$, estimated using Monte Carlo Markov Chain (MCMC) algorithm resulting in samples $\{\theta_i\}^{m}_{i=1}$. Given a set of observations $\mathcal{D}_{n} = (x_{1:n}, y_{1:n})$  at step $n$, then the prediction of the GP $y_{n+1} = \hat{y}(x_{n+1})$ at a new query point $x_{n+1}$, with kernel $k_i$ conditioned on the $i$-th hyperparameter sample $k_i = k(\cdot, \cdot|\theta_i)$ is normally distributed, $\hat{y}(x_{n+1})\sim \sum^{m}_{i=1} \mathcal{N}(\mu_i, \sigma^2_i|x_{n+1})$,
where:
\begin{equation}
    \label{eq:BO2}
    \mu_i(x_{n+1}) = k_i(x_{n+1}, X)K^{-1}_i y
\end{equation}
\begin{equation}
    \label{eq:BO3}
    \sigma^2_i(x_{n+1}) = k_i(x_{n+1}, x_{n+1}) - k_i(x_{n+1}, X)^T K^{-1}_i k_i(X, x_{n+1})
\end{equation}
where $k_i(x_{n+1}, X) = \left[k_i(x_{n+1}, x_j)\right]_{x_j \in X}$ and $K = \left[k_i(x_j, x_l)\right]_{x_j,x_l \in X} + \mathbf{I}\sigma^2_n$ being $\sigma^2_n$ the noise, with $\mu_i$ and $\sigma^2_i$ representing the prediction and uncertainty of the model at $x_{n+1}$.

The second one is an \textbf{acquisition function} $\alpha(x, \mathnormal{p(f)})$. It considers the predictive distribution for each point in $\mathcal{X}$ to select the next point at each iteration. The criteria used to perform this task is the Expected Improvement criterion (EI), which, given a set of observations $\mathcal{D}_{n} = (X, y)$ at step $n$ and considering the expectation over the predictive distribution, can be computed as:

\begin{equation}
    \label{eq:BO6}
    \begin{split}
        EI(x) = \mathbb{E}_{(\hat{y}|\mathcal{D}_n,\theta,x)} [max(0, \hat{y}(x) - \rho)]\\
        = \sum_{i=1}^{m} [(\mu_i(x) - \rho) \Phi(z_i) + \sigma_i(x) \phi(z_i)]
    \end{split}
\end{equation}

\noindent
where $\rho$ is an incumbent value, i.e. $\rho = max(y_{1:n})$, $\phi$ and $\Phi$ correspond to the probability and cumulative density functions respectively, and $z_i = (\rho - \mu_i(x))/\sigma_i(x)$. The pair $(\mu_i, \sigma^2_i)$ is the mean and variance of the predictive distribution. EI can be maximized by optimizing two main elements of Eq.~\ref{eq:BO6}, which are $(\mu_i(x) - \rho)$, for exploitation, and $\sigma_i(x)$, for exploration.

%%%%%%%%%%%%%%%%%%%%%%%%%%%%%%%%%%%%%%%%%%%%%%%%%%%%%%%%%%%%%%%%%%%%%%%%%%%%%%%%
\section{Grasp Evaluation}
\label{method}

Bringing the iterative process of BO to robotic grasping terms, in the first step of the optimization, the grasping process is fed with an initial set of samples $\mathcal{X}$ of the grasp pose.
%whether it is, for instance, the position of the fingers, the pose of the wrist, or the pose of the hand palm. 
For the selected set of samples, the grasping process takes place, updating the probabilistic model and the acquisition function with the grasp quality computed for each of the samples. In the remaining iterations, the pose of the hand that maximizes the acquisition function is computed and each time a new target pose is received, the hand is placed in the new location using position control, and force-controlled to grasp the object, and a new evaluation \newtext{of the function developed below} is carried out. We optimize $\{x,y,z\}$ coordinates of the hand palm based on the bounding box of the object plus the finger size in each direction. We also optimize the roll angle. Pitch and yaw are computed to align with the object center \cite{dlr197869}. It is important to mention that this 4D problem is higher than in previous research with 2D \cite{nogueira2016unscented} or 3D \cite{castanheira2018finding} parametrizations.
%below, the grasp pose was provided as the position of the hand palm reference in 3D, and an offset of its roll angle and the search space was delimited using a bounding box, which implementation is described in \cite{dlr197869}.

%dimensionality of the search space.

%We summarize the parametrization, including the bounding box depending on the length, width and height of the object and the length of the proximal phalanx of the robotic hand used.

The following section introduces all the processes that allow the method to be transferred from a limited simulation (Fig. \ref{fig:setups}A) to the real system (setups in Figs. \ref{fig:setups}B and \ref{fig:setups}C). 

\subsection{Reachability and collisions}
For finding potential grasp candidates, we freely sample the space around the object to retrieve potential grasping poses. However, 
%Although the exploration space is bounded depending on the object's shape, within these limits the grasping poses computed by BO are totally unconstrained. 
this means that some hand-palm poses and robot configurations can be unfeasible for different reasons.

As a first step to evaluate the grasping poses provided by the acquisition function of the BO, the reachability of such pose is checked by comparing it with a Capability Map \cite{porges2014reachability}, whose cells have an associated index representing the dexterity of the robot in that location. Once this check has been passed, we compute the robot inverse kinematics to provide a specific robot joint configuration.

Another type of unfeasible grasping pose occurs when the robot is in collision with either the object or the workbench. A possible collision with the workbench is handled by classifying the grasp as invalid, assigning a grasp quality equal to 0, thus moving the acquisition function away from poses similar to this one. For collisions between hand and object, an Approximation Reward ($AR$) is used to differentiate them from collisions with the workbench or from poses that do not even touch the object, thus guiding the search toward regions that are worth exploring. Inspired by the Collision Penalty ($CP$) in \cite{castanheira2018finding}, this $AR$ brings the optimization closer to the object while avoiding those poses in which object and hand collide, assigning a higher value to the lower level of penetration of the hand into the object, as follows:

\begin{equation}
\label{eq:CP2}
AR(n_j) =
\begin{cases}
e^{-\lambda n_j} , & \text{if } n_j > 0 \\
0, & \text{if } n_j = 0
\end{cases}
\end{equation}
where $\lambda$ acts as a decay rate constant and $n_j$ is the number of hand links colliding with the object in the pose sampled by BO. When there are no collisions between hand and object and $n_j = 0$, $AR$ is also 0. \newtext{This kind of heuristic does not need precise knowledge about object shape, and the pre-calculated collisions can be obtained using the robot model and a bounding box, which can be estimated in a real environment from a single point of view}.

\begin{figure}
    \centering
    \includegraphics[width=0.9\linewidth]{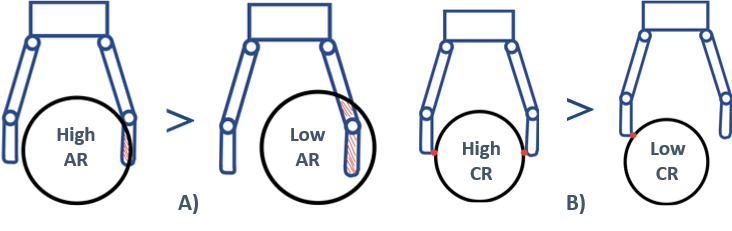}
    \caption{New metrics proposed: A) Approximation Reward in two grasps with collision, from higher $AR$ (left) to lower $AR$ (right). B) Contact Reward in two grasps with finger contact points, from higher $CR$ (left) to lower $CR$ (right).}
    \label{fig:CPyCR}
\end{figure}

\subsection{Absence of force closure}
In a typical grasp planning approach, when the grasp generated after the hand closes around the object does not lead to force closure, the grasp quality is equal to 0 to feed the optimization. However, taking as inspiration the $AR$, a Contact Reward ($CR$) has been implemented to improve the method's convergence. This $CR$ is defined as:
\begin{equation}
\label{eq:CRa }   
CR(n_c) = 1- e^{-\lambda n_c}
\end{equation}
where $n_c$ is the number of contact points at the hand's fingertips and $\lambda$ has the same purpose as in Eq.~\ref{eq:CP2}. In this way, the acquisition function is directed towards positions where there is a higher number of fingertip contacts with the object.

\subsection{Grasp quality computation}
When an iteration results in a grasp with force closure, the quality of the grasp is computed. Several works \cite{roa2015grasp,leon2014characterization} have concluded that the use of a combined metric is advantageous for finding an optimal realistic grasp. Based on these works, the following metrics were studied: to consider the hand configuration, the Uniformity of the transformation in the velocity domain from the finger joints to the object is computed using the condition number of the Jacobian matrix ($Q_{uni}$) \cite{salisbury1982articulated}; the Grasp Isotropy Index ($Q_{iso}$) \cite{kim2001optimal} is computed from the singular values of the Grasp matrix to indicate an isotropic contribution of contact forces to the applied wrench; further analysis of the contact forces is performed using the wrenches that the grasp can resist, known as Grasp Wrench Space (GWS), by considering the Largest-minimum resisted wrench ($Q_\epsilon$) \cite{ferrari1992planning} and the Volume of this GWS ($Q_v$) \cite{li1988task}. From experiments in sec. \ref{simstudy} and \ref{exp0}, weights were assigned to each metric, obtaining the combined metric in Eq.~\ref{Qm}, where $\sum^4_{i=1} w_i = 1$. Weights can also be optimized by other methods making $Q_m$ more robust.

\begin{equation}
    Q_m = w_1 * Q_{iso} + w_2 * Q_{\epsilon} + w_3 * Q_{v} + w_4 * Q_{uni}
    \label{Qm}
\end{equation}

\noindent
$Q_m$ also aims to compensate metrics with worse convergence with those with better convergence and to compensate those more influenced by the uncertainty of the real environment. The resulting evaluation function is of the form:

\begin{equation}
\label{eq:Qgp}   
y_{gr} = Q_c * (Q_m + \alpha (AR + CR))
\end{equation}

\noindent
where $Q_c$ is a binary index indicating whether the pose is feasible or not. The $\alpha$ coefficient has been added to regulate the range of values spanned by $AR$ and $CR$, depending on the magnitude of the values obtained when calculating the grasp quality metrics. In this way, a higher outcome is always given for a pose in which the object is successfully grasped. Note that both the $AR$ and $CR$ are intended to guide the acquisition function without hindering the approximation to a robust grasp.

%%%%%%%%%%%%%%%%%%%%%%%%%%%%%%%%%%%%%%%%%%%%%%%%%%%%%%%%%%%%%%%%%%%%%%%%%%%%%%%%
\section{Experimental results}

\begin{figure}
\centering
 \includegraphics[width=0.9\linewidth]{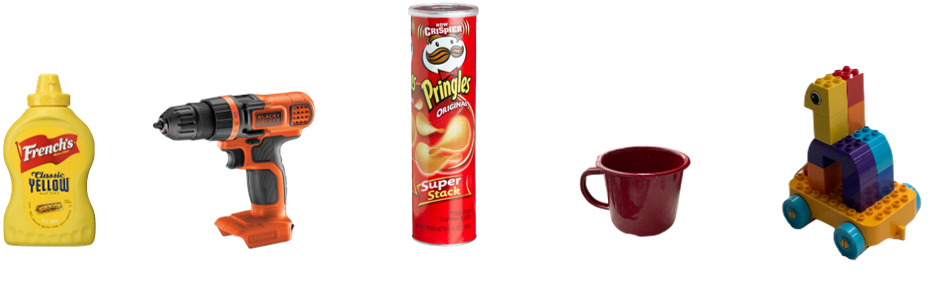}
\caption{Objects from the YCB set used for experimentation, both in simulation and in the real environment. They show variety in shapes, weights, sizes, symmetry or the position of the center of mass.} 
\label{fig:ycb}
\end{figure}

In this section, we present the experiments conducted in simulation and with the real robot. The simulation was implemented using PyBullet, and the objects considered (Fig.~\ref{fig:ycb}) were selected from the publicly available YCB object set \cite{7254318}.

\subsection{Ablation study}
\label{simstudy}
\begin{figure}
\centering
 \includegraphics[width=0.9\linewidth]{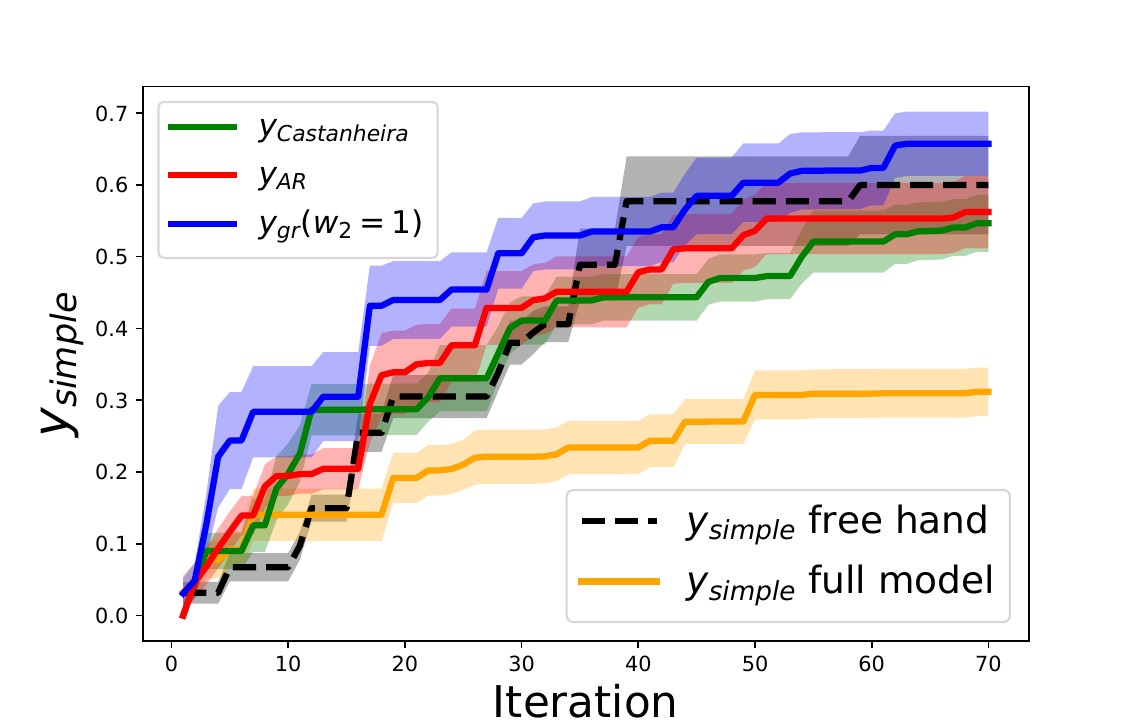}
\caption{Comparison of the evolution of the $y_{simple}$ outcome, obtained using different implementations of the grasp evaluation function.}
\label{fig:graf2}
\end{figure}

For establishing a baseline for comparison, we use the free-hand simulation from \cite{herrera2022optimizacion}, tested in the CLASH hand with $w_2=1$. The baseline is presented with a dashed line in Fig. \ref{fig:graf2} and in Table \ref{tab:tab1.1}, showing BO's ease in finding feasible grasps with great outcome values. The corresponding evaluation function has the form presented in Eq.~\ref{eq:exp2}, where $Q_f$ is a binary index indicating the force closure condition in this experimental case.
\begin{equation}
\label{eq:exp2}   
y_{simple} = Q_{c} * Q_{f} * Q_m(w_2 = 1)
\end{equation}

The following experiments show a comparison between the results obtained using different implementations of the grasp evaluation function on the complete robot model simulation, shown in Fig. \ref{fig:setups}B. \newtext{As in the previous case, we use $w_2=1$, with 20 initial samples and 50 BO iterations to optimize with each evaluation function. The plots include a comparison of the outcomes obtained in Eq.~\ref{eq:exp2}, as a common metric, with the grasps found by each evaluation function in 10 optimizations, and the shaded area represents the 95\% confidence interval.}

\begin{table}
\centering
\caption{Ablation study. Mean and maximum $y_{simple}$ outcomes obtained grasping a mustard bottle with different grasp evaluation functions. FH: free hand; FM: full model.}
\label{tab:tab1.1}
\begin{tabular}{ll|ll|l|l|l|}
\cline{3-7}
\multicolumn{2}{l|}{\multirow{2}{*}{}}                                                                                   & \multicolumn{2}{l|}{\textbf{$y_{simple}$}}                                                                                                         & \multirow{2}{*}{\textbf{\begin{tabular}[c]{@{}l@{}}Cas.\cite{castanheira2018finding}\\  FM\end{tabular}}} & \multirow{2}{*}{\textbf{\begin{tabular}[c]{@{}l@{}}$y_{AR}$\\ FM\end{tabular}}} & \multirow{2}{*}{\textbf{\begin{tabular}[c]{@{}l@{}}$y_{gr}$\\ FM\end{tabular}}} \\ \cline{3-4}
\multicolumn{2}{l|}{}                                                                                                    & \multicolumn{1}{l|}{\textbf{FH}} & \textbf{FM} &                                    &                                        &                                       \\ \hline
\multicolumn{1}{|l|}{\multirow{2}{*}{\textbf{\begin{tabular}[c]{@{}l@{}}Mustard\\ bottle\end{tabular}}}} & \textbf{Mean} & \multicolumn{1}{l|}{0.599}                                                        & 0.318                                                          & 0.546                              & 0.562                                  & 0.657                                 \\ \cline{2-7} 
\multicolumn{1}{|l|}{}                                                                                   & \textbf{Max.} & \multicolumn{1}{l|}{0.967}                                                        & 0.518                                                          & 0.815                              & 0.901                                  & 0.935                                 \\ \hline
\end{tabular}
\end{table}

\begin{table}
\centering
\caption{Full model simulation. Mean and maximum outcomes obtained by grasping different objects using $w_2$ = 1 in $y_{gr}$.}
\label{tab:tab1.2}
\begin{tabular}{l|l|l|l|}
\cline{2-4}
                                    & \textbf{Mustard bottle} & \textbf{Mug} & \textbf{Power drill} \\ \hline
\multicolumn{1}{|l|}{\textbf{Mean}} & 0.6572           & 0.6062       & 0.4964               \\ \hline
\multicolumn{1}{|l|}{\textbf{Max.}} & 0.9345           & 0.8713       & 0.6246               \\ \hline
\end{tabular}
\end{table}

When adding the arm and the workbench in the simulation, the obtained metric values are much lower than in the case of the free-hand simulation, as observed in Table \ref{tab:tab1.1} and the orange line in Fig.~\ref{fig:graf2}. The reachability of the grasping pose and the collisions detected when including the arm and the workbench result in a higher number of grasping poses evaluated as 0. 
We also added the AR value to the evaluation function, leading to Eq.~\ref{eq:exp3}. Throughout all the experimentation, the tuning parameter was set to $\lambda = 0.1$, and a coefficient $\alpha = 0.1$ was used, resulting in $AR$ values between 0 and 0.1, being substantially lower than those provided by the grasp metric, which has been normalized between 0 and 1.

\begin{equation}
\label{eq:exp3}   
y_{AR} = Q_c * Q_{f} * (Q_m(w_2 = 1) + \alpha AR)
\end{equation}

Table \ref{tab:tab1.1} shows how both mean and maximum outcomes increase with the addition of $AR$, and in Fig. \ref{fig:graf2} we can see how active learning reaches much higher values already in the first iterations. The comparison between the results obtained by applying the $CP$ \cite{castanheira2018finding} and the $AR$ proposed in this work demonstrates how assigning a higher value to $AR$ the lower the degree of collision with the object drives the acquisition function towards near-object poses, improving convergence while preventing collisions.

In the next step of the experimentation, we added $CR$, resulting in Eq.~\ref{eq:Qgp}, again with $w_2=1$. The results of Fig.~\ref{fig:graf2} and Table \ref{tab:tab1.1} show the greater convergence of the method when including this coefficient. Table~\ref{tab:tab1.2} shows the results obtained on more complex objects, with $y_{gr}$ greater than $0.5$.

%%%%%%%%%%%%%%%%%%%%%%%%%%%%%%%%%%%%%

To continue with the full model simulation experiments, the performance of the method using each metric was tested. Table \ref{tab:tab4} shows the time for each metric to achieve a $y_{gr}$ outcome greater than $0.5$, and the number of times that this minimum outcome was not reached. \newtext{Considering metrics normalization, we choose this threshold based on the literature \cite{leon2014characterization}, providing a logical division between above and below-average performances and proving to be correlated with a reasonable success rate. It could be increased to provide more demanding applications. However, complementing this approach with experimental validation is key for a meaningful evaluation.} We evaluated the grasps with the function provided in Eq.~\ref{eq:Qgp}, assigning a weight of 1 to the corresponding grasp metric ($Q_{iso}, Q_\epsilon, Q_v, Q_{uni}$). The same number of iterations as in the previous experiments was used.

\begin{table}
\centering
\caption{Full model simulation: Convergence of each metric until $y_{gr} >$ 0.5.}
\label{tab:tab4}
\begin{tabular}{ll|l|l|l|l|}
\cline{3-6}
                                                                                                       &                                    & \textbf{$w_1$ = 1} & \textbf{$w_2$ = 1} & \textbf{$w_3$ = 1} & \textbf{$w_4$ = 1} \\ \hline
\multicolumn{1}{|l|}{\multirow{2}{*}{{\textbf{\begin{tabular}[c]{@{}l@{}}Mustard \\ bottle\end{tabular}}}}}                                                & \textbf{time (s)}                & 12.52              & 31.00              & 55.46              & 18.77              \\ \cline{2-6} 
\multicolumn{1}{|l|}{}                                                                                 & \textbf{$y_{gr}$ \textless 0.5} & 0/10               & 3/10               & 5/10               & 0/10               \\ \hline
\multicolumn{1}{|l|}{\multirow{2}{*}{\textbf{Mug}}}                                                    & \textbf{time (s)}                & 21.31              & 32.93              & 46.13              & 14.94              \\ \cline{2-6} 
\multicolumn{1}{|l|}{}                                                                                 & \textbf{$y_{gr}$ \textless 0.5} & 0/10               & 4/10               & 5/10               & 0/10               \\ \hline
\multicolumn{1}{|l|}{\multirow{2}{*}{\textbf{\begin{tabular}[c]{@{}l@{}}Power \\ drill\end{tabular}}}} & \textbf{time (s)}                & 27.16              & 38.95              & 38.15              & 12.02              \\ \cline{2-6} 
\multicolumn{1}{|l|}{}                                                                                 & \textbf{$y_{gr}$ \textless 0.5} & 3/10               & 6/10               & 6/10               & 0/10               \\ \hline
\end{tabular}
\end{table}

\begin{figure}
\centering
 \includegraphics[width=1\linewidth]{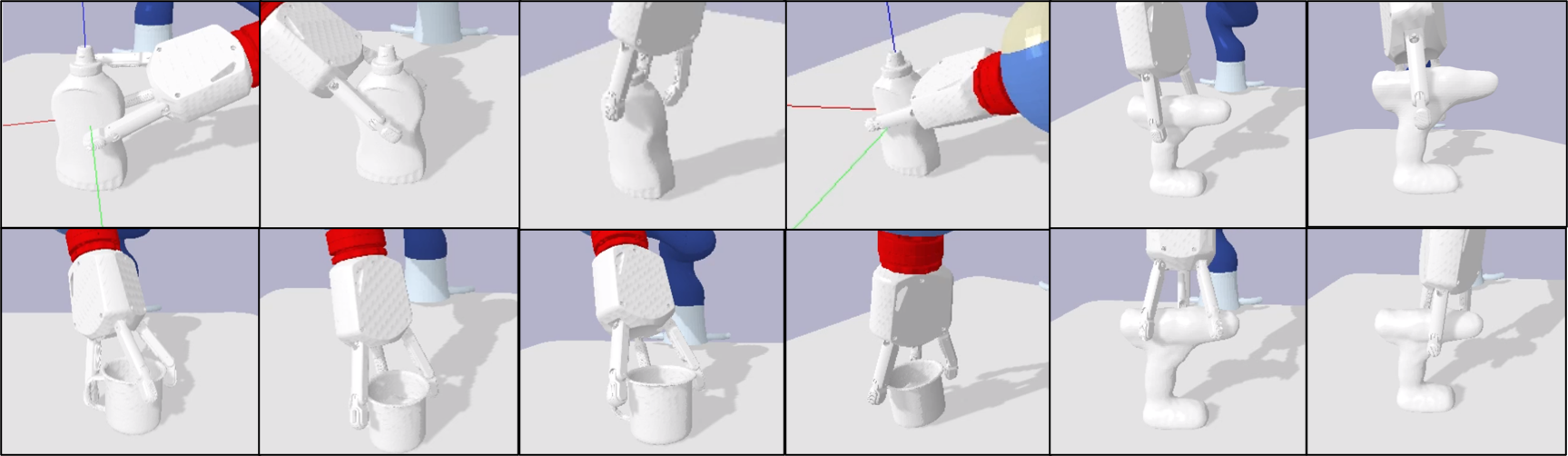}
\caption{Some of the best grasps obtained with each metric in simulation, grasping the mustard bottle, the mug, and the power drill. From left to right: $Q_{iso}$ ($w_1$ = 1), $Q_\epsilon$ ($w_2$ = 1), $Q_v$ ($w_3$ = 1) and $Q_{uni}$ ($w_4$ = 1).} 
\label{fig:simresults1}
\end{figure}

The average convergence time obtained with $Q_{iso}$ and $Q_{uni}$ is much lower than the time obtained when using $Q_\epsilon$ and $Q_v$. Note that these last two metrics did not reach the minimum $y_{gr}$ outcome a greater number of times, while $Q_{uni}$ is the only metric that reached this minimum value in the totality of the experiments. On the other hand, $Q_v$ was the metric with the worst performance in obtaining an optimal grasp on each object. $Q_{iso}$ and $Q_{uni}$ are ratios of the minimum and maximum singular values of their respective matrices and are therefore unitless. When normalized, their range of values will be lower and closer to 1 if the differences between the maximum and minimum singular values on each case remain similar. On the other hand, $Q_\epsilon$ and $Q_v$ are absolute measures from the wrench space. Their variation can be more accentuated because of the influence of small disturbances in the position or direction of the forces exerted on the object. Fig. \ref{fig:simresults1} shows some of the best grasps obtained by the optimization of Eq.~\ref{eq:Qgp} during the described experiments, grasping each considered object.

%########################################################################################################

\subsection{Real-world setup}
\label{setup}

Our robot setup, presented in Fig. \ref{fig:setups}C consists of a 7 degrees of freedom LWR arm \cite{hirzinger2008light} and a three-fingered CLASH hand \cite{friedl2020clash}, with variable stiffness in the fingers and including a grid of 4x4 haptic sensors XELA uSkin Patch \cite{XELA} on the fingertips, measuring force in three axes. Together with the finger coupling, CLASH variable stiffness allows greater adaptability in grasps with different types of objects, in which a lower tendon pretension can prevent the slippage of a curved object or deformation of objects of low consistency when pressure is exerted. The contact point analysis requires the computation of the pose of each sensor using the kinematics of the arm-hand assembly. For this purpose, the hand has been considered fully actuated, which does not correspond to reality, but sill provides a very good estimation of the finger poses. Carrying out the real experimentation with an under-actuated hand adds more error to the calculations performed to evaluate the grasp. The tendon-driven fingers make it difficult to obtain an accurate joint position, which translates into another source of uncertainty that BO has to handle. The CLASH hand provides interfaces to control the finger's stiffness and joint positions independently, and the LWR arm provides interfaces for both Cartesian and joint impedance control. These controllers run on a Linux real-time machine, while other software components, including BayesOpt and the Pybullet simulation for collision avoidance, are executed in a separate Linux PC with an Intel Xeon E5-1630 v4 CPU @ 3.70GHz and an NVIDIA Quadro K620 GPU. 

One restriction of performing movements on the real arm is that most grasping poses result in collisions when the hand approaches the target directly. To solve this type of collision a pre-grasp position is established, several centimeters away from the target position in the local Z-axis of the hand, from which the hand approaches the object safely every iteration. Additional information on the implementation problems addressed can be found in \cite{dlr197869}.

%########################################################################################################

\begin{figure*}
\centering
 \subfloat[]{\includegraphics[width=0.32\linewidth]{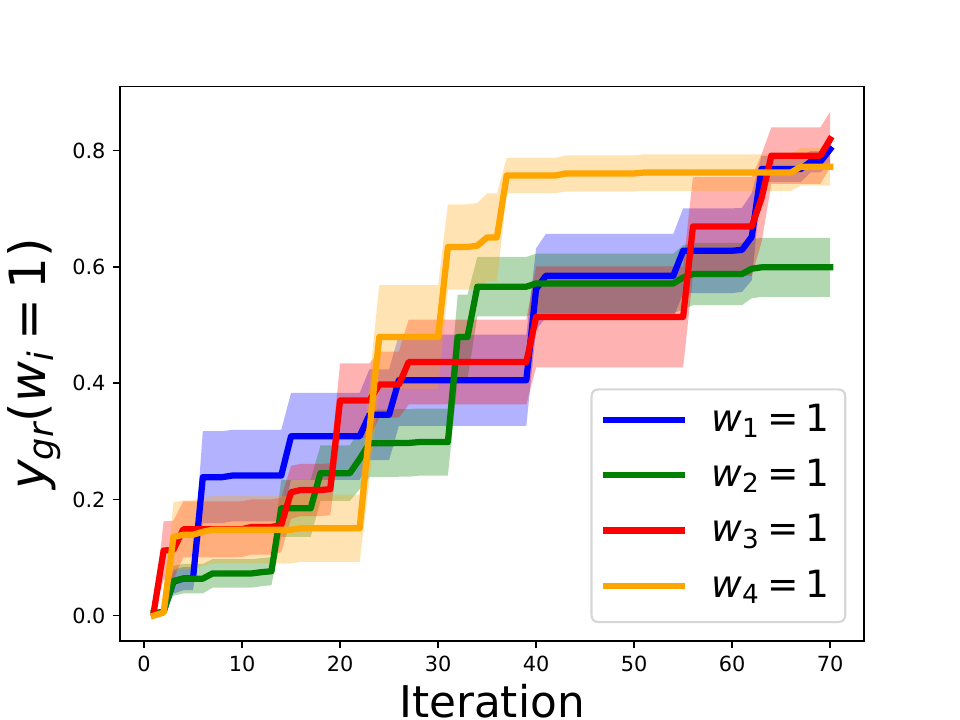}%
 \label{grafexp1A}}
 \subfloat[]{\includegraphics[width=0.32\linewidth]{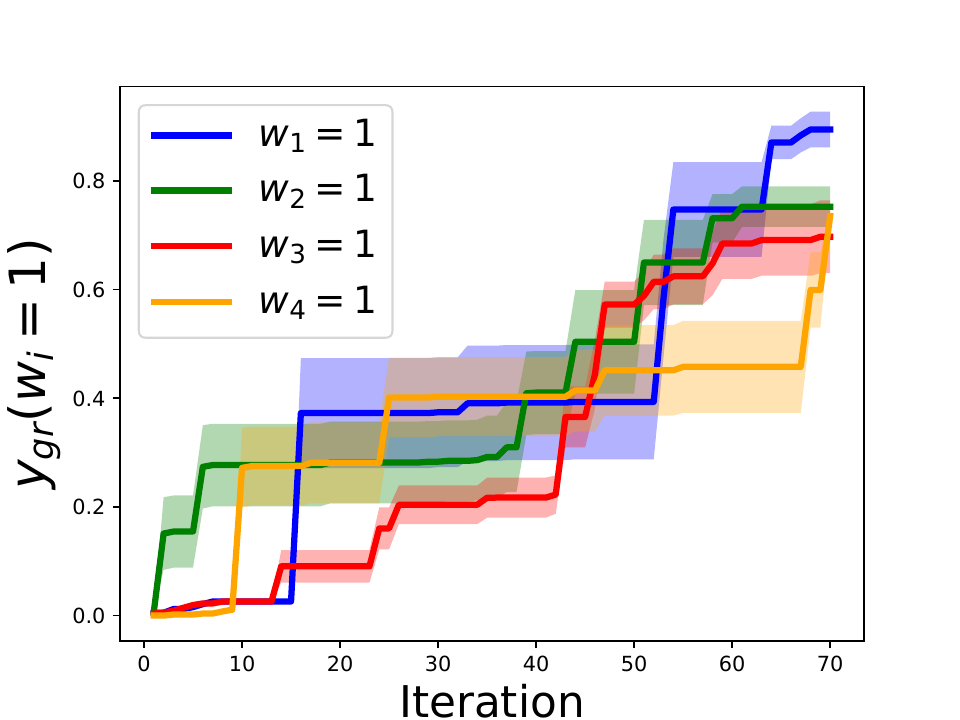}%
 \label{grafexp1B}}
 \subfloat[]{\includegraphics[width=0.32\linewidth]{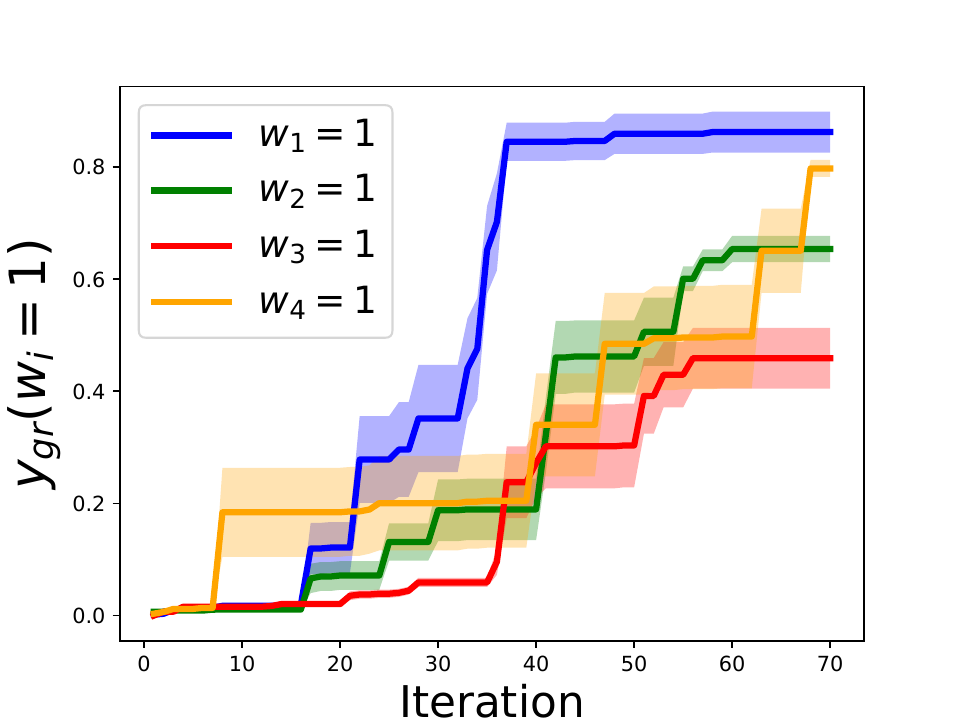}%
 \label{grafexp1C}}
\caption{Comparison of the evolution of the grasp quality, to test optimal results in the real setup. Each line represents the outcome of Eq.~\ref{eq:Qgp}, assigning a value of 1 to the corresponding weight of each metric and grasping different objects: a) Pringles can. b) Mug. c) Lego toy.} 
\label{fig:grafexp1}
\end{figure*}

\subsection{Simulation vs. real robot}
\label{exp0}

To verify the portability of the method to the real robot, in this experiment different grasping poses were obtained in simulation using each of the implemented metrics and three different objects, and then tested on the real robot. The number of successful grasps in 5 optimizations, using the same number of iterations as in the previous experiments, was obtained for each metric and object, \newtext{to compare the metrics' effectivenes}. The outcomes have been plotted in Fig.~\ref{fig:grafexp1} with their 95\% confidence intervals, \newtext{representing the learning differences of each metric}. \newtext{To evaluate grasp success}, the object was lifted several centimeters from the table and held in that position for a period of 5 seconds. A binary score (success or fail) was recorded for each optimal grasp pose. A grasp execution was considered a failure if the object fell or slipped during its lift motion, touching the workbench. The results obtained in this experiment are shown in Table~\ref{tab:tab5}.

\begin{table}
\centering
\caption{Mean outcomes and successful grasps for each metric, optimized in simulation and tested in the real robot.}
\label{tab:tab5}
\begin{tabular}{ll|l|l|l|l|}
\cline{3-6}
                                                         &                                                                  & \textbf{$w_1$ = 1} & \textbf{$w_2$ = 1} & \textbf{$w_3$ = 1} & \textbf{$w_4$ = 1} \\ \hline
\multicolumn{1}{|l|}{\multirow{2}{*}{\textbf{Pringles}}} & \textbf{\begin{tabular}[c]{@{}l@{}}Mean \\ outcome\end{tabular}} & 0.8023             & 0.6991             & 0.8913             & 0.7722             \\ \cline{2-6} 
\multicolumn{1}{|l|}{}                                   & \textbf{Success}                                                 & 3/5                & 3/5                & 1/5                & 3/5                \\ \hline
\multicolumn{1}{|l|}{\multirow{2}{*}{\textbf{Mug}}}      & \textbf{\begin{tabular}[c]{@{}l@{}}Mean \\ outcome\end{tabular}} & 0.8943             & 0.7526             & 0.6972             & 0.7354             \\ \cline{2-6} 
\multicolumn{1}{|l|}{}                                   & \textbf{Success}                                                 & 2/5                & 4/5                & 3/5                & 4/5                \\ \hline
\multicolumn{1}{|l|}{\multirow{2}{*}{\textbf{Lego}}}     & \textbf{\begin{tabular}[c]{@{}l@{}}Mean \\ outcome\end{tabular}} & 0.8623             & 0.6537             & 0.4590             & 0.7974             \\ \cline{2-6} 
\multicolumn{1}{|l|}{}                                   & \textbf{Success}                                                 & 3/5                & 2/5                & 1/5                & 2/5                \\ \hline
\multicolumn{2}{|l|}{\textbf{Total Success}}                                                                                & 8/15               & 9/15               & 5/15               & 9/15               \\ \hline
\end{tabular}
\end{table}

\begin{figure*}
\centering
 \includegraphics[width=0.85\linewidth]{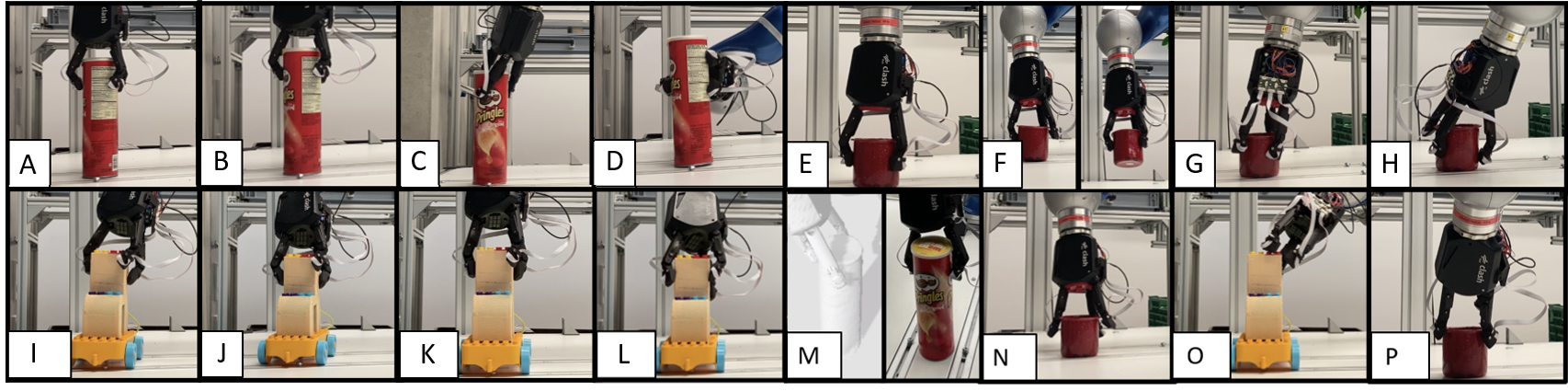}
\caption{Grasps optimized in simulation with each metric and tested on the real robot. Figures A-D correspond to successful grasps holding the Pringles can, using $Q_{iso}, Q_{\epsilon}, Q_v, Q_{uni}$, in this order. Figures E-H and I-L correspond to successful grasps holding the mug and Pringles can respectively using the same metrics order. Figures M-P correspond to failed grasps due to slippage or mislocation.}
\label{fig:exp0agarres}
\end{figure*}

We can observe mean $y_{gr}$ outcomes greater than 0.5 in most of the cases. This is especially interesting in the case of $Q_v$ ($w_3$ = 1), which only obtained 33.3\% effective grasps in the total evaluated. Both $Q_v$ and $Q_\epsilon$ are metrics based on GWS properties, which makes them very sensitive to positional errors \cite{weisz2012pose} becoming less accurate when working with sparse and noisy sensor information from the real robot. However, $Q_\epsilon$ is more effective, since it focuses on the maximum force that the grasp can resist and not on the whole wrench space. As seen in Fig. \ref{fig:grafexp1}, the more complex shape of the object slows down the convergence. However, in most of the cases, the metrics reach an average $y_{gr}$ outcome greater than $0.5$, and the percentage of effective grasps is 65\% of the 20 performed per object. In this figure, the blue line corresponding to $Q_{iso}$ ($w_1$ = 1) reaches the highest outcomes, however, only 2 of the 5 grasps performed succeeded in lifting the mug (Fig. \ref{grafexp1B}). This indicates a more symmetrical distribution of forces and torques around the object, as in Figs. \ref{fig:exp0agarres}E and \ref{fig:exp0agarres}G, is not always a sufficient condition to obtain an optimal grasp when faced with the uncertainty introduced by positional errors and an unknown friction coefficient. Although Fig. \ref{grafexp1A} shows very similar learning in all metrics when grasping the Pringles can, in the optimal grasps shown in Fig. \ref{fig:exp0agarres} we can observe certain trends related to the computation of each metric. While the other metrics focused on exploring the top, $Q_{uni}$ ($w_4$ = 1) found side grasps in which the hand grasped the can very robustly (Fig.  \ref{fig:exp0agarres}D). Since CLASH only has tactile sensors on the fingertips, when the Pringles are grasped from above in a sufficiently symmetrical configuration of the fingers, the direction of the contact forces results in a stable grasp according to $Q_v$, $Q_{iso}$ and $Q_\epsilon$. However, in our setup, a grasp with the hand wrapped around the can has similar contact points to a grasp touching the can from the side. Considering contact points only on the distal phalanges, and not in the proximals too, causes information to be lost in the analysis. Since objects are not static in the real world, Figs. \ref{fig:exp0agarres}M and \ref{fig:exp0agarres}N show grasps in which the object slipped from the hand when pressure was exerted. Figs. \ref{fig:exp0agarres}O and \ref{fig:exp0agarres}P show cases where the hand was located far away from the object and the force closure calculation was wrong. On the other hand, Fig. \ref{fig:exp0agarres}F shows a case in which the object's movement when exerting pressure favored the grasp since the mug was pushed inside the hand. Fig. \ref{fig:exp0agarres}H is another example of how the result obtained by $Q_{uni}$ wrapped around the mug to a greater extent. In the case of the Lego toy, two types of successful grasps can be observed: the ones in which one or more of the fingers were stuck on the toy's head, such as those shown in Figs. \ref{fig:exp0agarres}J and \ref{fig:exp0agarres}L, and those grasps in which the fingers made contact on the edges of the object as shown in Figs. \ref{fig:exp0agarres}I and \ref{fig:exp0agarres}K. It can also be seen how variable stiffness has increased adaptability to curved surfaces, as in Figs. \ref{fig:exp0agarres}C and \ref{fig:exp0agarres}H.

\subsection{Optimization in real environment}
\label{exp1}
\begin{figure}
\centering
 \includegraphics[width=0.8\linewidth]{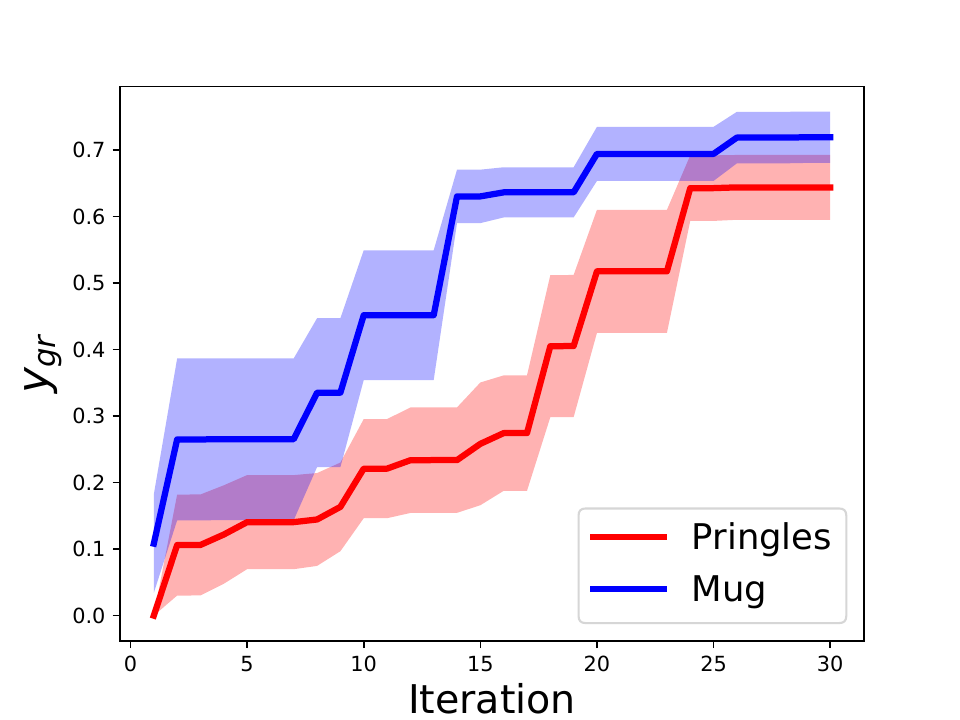}
\caption{Comparison of the evolution of the $y_{gr}$ outcome, optimizing the grasp directly in the real environment.} 
\label{fig:grafexp2outcome}
\end{figure}

In the following experiments, 5 initial samples and 25 BO iterations were performed directly on the real robot, using as evaluation function Eq.~\ref{eq:Qgp}. The weights in Eq.~\ref{Qm} were computed as the successful grasps obtained with each corresponding metric in sec. \ref{exp0}, divided by the total number of successful grasps. Collisions and reachability were computed in simulation but the quality of the grasp was obtained from the sensors and measurements provided by the robotic system. Table \ref{tab:tab6} shows the number of successful grasps in 5 optimizations and the outcomes obtained, which were also represented in Fig. \ref{fig:grafexp2outcome} with their 95\% confidence intervals.

\begin{table}
\centering
\caption{Optimization and testing in the real environment. Mean $y_{gr}$ outcomes and number of successful grasps.}
\label{tab:tab6}
\begin{tabular}{ll|l|l|l|l|}
\cline{3-6}
                            &                            & \textbf{\begin{tabular}[c]{@{}l@{}}Mean \\ outcome\end{tabular}} & \textbf{\begin{tabular}[c]{@{}l@{}}Best \\ outcome\end{tabular}} & \textbf{\begin{tabular}[c]{@{}l@{}}Success \\ over 5\end{tabular}} & \textbf{\begin{tabular}[c]{@{}l@{}}Feasible \\ grasp found\end{tabular}} \\ \hline
\multicolumn{2}{|l|}{\multirow{2}{*}{\textbf{Pringles}}} & \multirow{2}{*}{0.6436}                                          & \multirow{2}{*}{0.8423}                                          & \multirow{2}{*}{3/5}                                              & \multirow{2}{*}{5/5}                                                     \\
\multicolumn{2}{|l|}{}                                   &                                                                  &                                                                  &                                                                   &                                                                          \\ \hline
\multicolumn{2}{|l|}{\multirow{2}{*}{\textbf{Mug}}}      & \multirow{2}{*}{0.7189}                                          & \multirow{2}{*}{0.8579}                                          & \multirow{2}{*}{4/5}                                              & \multirow{2}{*}{5/5}                                                     \\
\multicolumn{2}{|l|}{}                                   &                                                                  &                                                                  &                                                                   &                                                                          \\ \hline
\end{tabular}
\end{table}

\begin{figure}
\centering
 \includegraphics[width=0.9\linewidth]{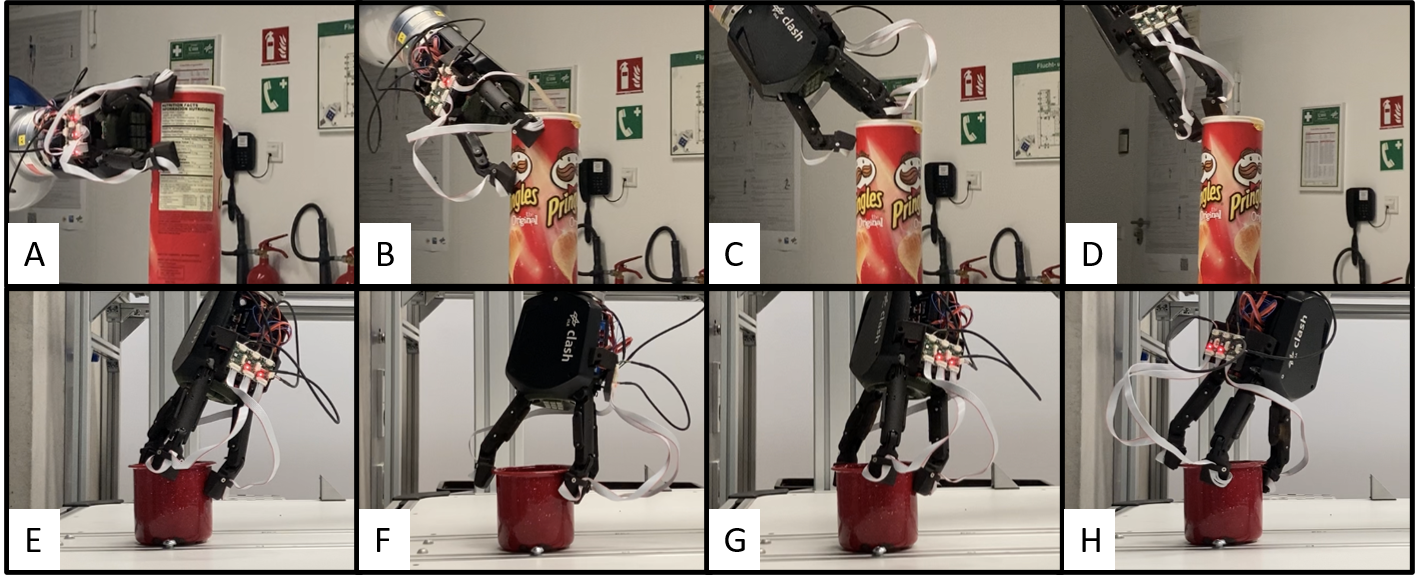}
\caption{Grasps optimizing $y_grasp$ directly in the real robot. Figures A and B show successful grasp in different locations holding the Pringles can. Figures C and D show failed grasps due to errors in force closure estimation, as well as figure H for the case of the mug. Figures E-G show successful grasp in the mug, varying the roll angle.}
\label{fig:exp2agarres}
\end{figure}

In the first optimizations, the importance of the initial samples to conform a well-informed prior can be observed, especially when a reduced number of iterations is considered. In the case of the Pringles can, when a high outcome side grasp was found in the initial samples, the search was focused on that area, more related to greater outcomes in $Q_{uni}$, resulting in a grasp like the one in Fig. \ref{fig:exp2agarres}A. High outcomes in the upper area of the Pringles focused the rest of the search on grasps related to $Q_\epsilon$ or $Q_{iso}$, like the one in Fig. \ref{fig:exp2agarres}B. Although the search space is unconstrained, addressing the learning with heuristics focused on improving the prior can be a good solution for the convergence of the method in areas of interest.

When working with an underactuated hand of variable stiffness, conventional analytical metrics become more inaccurate as they do not take into account positional errors due to differences in tendon pretension. The force closure computation proved to be insufficient to deal with these errors. Several false good grasps were observed throughout the iterations, two of which are shown in Figs. \ref{fig:exp2agarres}C and \ref{fig:exp2agarres}D. In general, the use of the GWS to classify the grasp has been proven to have a considerable error due to the uncertainty of the measurements taken for its calculation. Except for $Q_\epsilon$, none of the metrics considered guarantees the force closure condition in the grasp. Since $Q_{iso}$ and $Q_{uni}$ by their definition give rise to more symmetric contact point distributions surrounding the object, $Q_v$ is the one with the highest risk of obtaining high values for ineffective grasps such as in Fig. \ref{fig:exp2agarres}H. Even so, Bayesian optimization has been shown to deal positively with these difficulties, achieving a 70\% success rate in the optimization of the grasp performed completely on the real robot, as indicated in Table \ref{tab:tab6}. Note also that although in 30\% of the cases the optimal grasp was not successful, in all cases BayesOpt found other feasible and robust grasps assigned with lower metric values. However, the advantages and disadvantages of each of the metrics have been slightly compensated by using $Q_m$ to observe a greater variety in the obtained grasps.

Fig. \ref{fig:grafexp2outcome} shows faster learning, with higher outcomes for the mug case. The size of the Pringles can, especially in its Z-axis, reduces the convergence of the method but allows obtaining lateral clamp-like grasps. The exploration of the mug gave results that were as successful as those observed in Figs. \ref{fig:exp2agarres}E to G, with more significant variations in the roll angle.

Although the results obtained in \cite{joshi2020robotic} give evidence of the effectiveness of their method, they are subject to a prior learning several orders of magnitude higher than the number of trials used in this work to obtain a robust grasp. Moreover, in our case the learning is performed online for any new object. On the other hand, the grasp planner proposed in \cite{mahler2017dex} presents very competitive planning times of less than one second, but which, again, need prior training with large datasets directly affecting the accuracy of the method. The convergence times presented in table \ref{tab:tab4} depend on both software and hardware experimental conditions and still do not exceed one minute of learning in any case, improving also the results obtained with other model-free approaches such as the one presented in \cite{mahler2015gp}.

%%%%%%%%%%%%%%%%%%%%%%%%%%%%%%%%%%%%%%%%%%%%%%%%%%%%%%%%%%%%%%%%%%%%%%%%%%%%%%%%
\section{Conclusions}
%Conclusions
This work has presented a practical application of BO to the tactile exploration of objects in real environments. It has also shown the effectiveness of BO for grasping diverse and previously unknown objects in a robust and secure manner, using tactile exploration with real-life noise conditions. By conducting comprehensive experiments in simulated and real robotic environments, we have gained valuable insights into the adaptability and performance of this approach in a real setting, accounting for uncertainties and technical complexities.

The use of BO allows the addition of different metrics and heuristics to the grasp evaluation function to deal with diverse environments, without compromising the number of iterations needed. Among these heuristics, an Approximation Reward $AR$, a variation to the classical Collision Penalty $CP$, has been proposed to force the search algorithm to move away from possible collision, and a Contact Reward $CR$ has also been included to bring the search closer to more feasible grasping configurations, thus accelerating convergence.

Our consideration of various grasp quality metrics has shed light on their practical utility when applied to a real system, and how they affect the results obtained in BO, which worked significantly well when using noisy measurements to evaluate each grasp trial.

%Our consideration of various grasp quality metrics has shed light on their practical utility when applied to a real system, and how their behavior affects the results obtained in BO.

The  use of a variable stiffness hand has been beneficial, since its great adaptability allows to avoid slipping and to grasp curved surfaces robustly. 
Our next steps include the consideration of metrics and a grasp parameterization that better exploits the characteristics of a variable stiffness hand.

%%%%%%%%%%%%%%%%%%%%%%%%%%%%%%%%%%%%%%%%%%%%%%%%%%%%%%%%%%%%%%%%%%%%%%%%%%%%%%%%
\section*{Acknowledgment}
\label{sec:acknowledgement}
This work was partially supported by the EU Horizon Europe research and innovation program under Grant 101070136 (Intelliman), Spanish Government (PID2021-125209OB-I00 and PID2020-114819GB-I00, CaRo), and the Aragon Government (DGA-T45\textunderscore23R).

%endispell

\bibliographystyle{./style/IEEEtran}
\bibliography{bibliography}

\end{document}